\documentclass{article}
\usepackage{stfloats}
\usepackage[T1]{fontenc}
\usepackage{amssymb,amsmath}
\usepackage{txfonts}
\usepackage{microtype}
\usepackage{xspace}
\xspaceaddexceptions{\%}

\usepackage{paralist}

\usepackage{graphicx}
\usepackage{subfig} 

\usepackage{natbib}

\usepackage{algorithm}
\usepackage{algorithmic}

\usepackage[hyphens]{url}
\usepackage{hyperref}

\usepackage{mlp2021}
\mlptitlerunning{MLP Coursework 4 -- Final Report (\groupNumber)}
\usepackage{multirow}

\def\projectTitle{Generative Artisan}

\title{\projectTitle: A Semantic-Aware and Controllable CLIPstyler}
\date{}
\author{Zhenling Yang,  Huacheng Song,  Qiunan Wu\\ The University of Edinburgh\\ s1862671@ed.ac.uk}
\begin{document} 
\twocolumn[{
\maketitle
 \begin{figure}[H]
 \hsize=\textwidth
 \begin{center}
 \includegraphics[width=\textwidth]{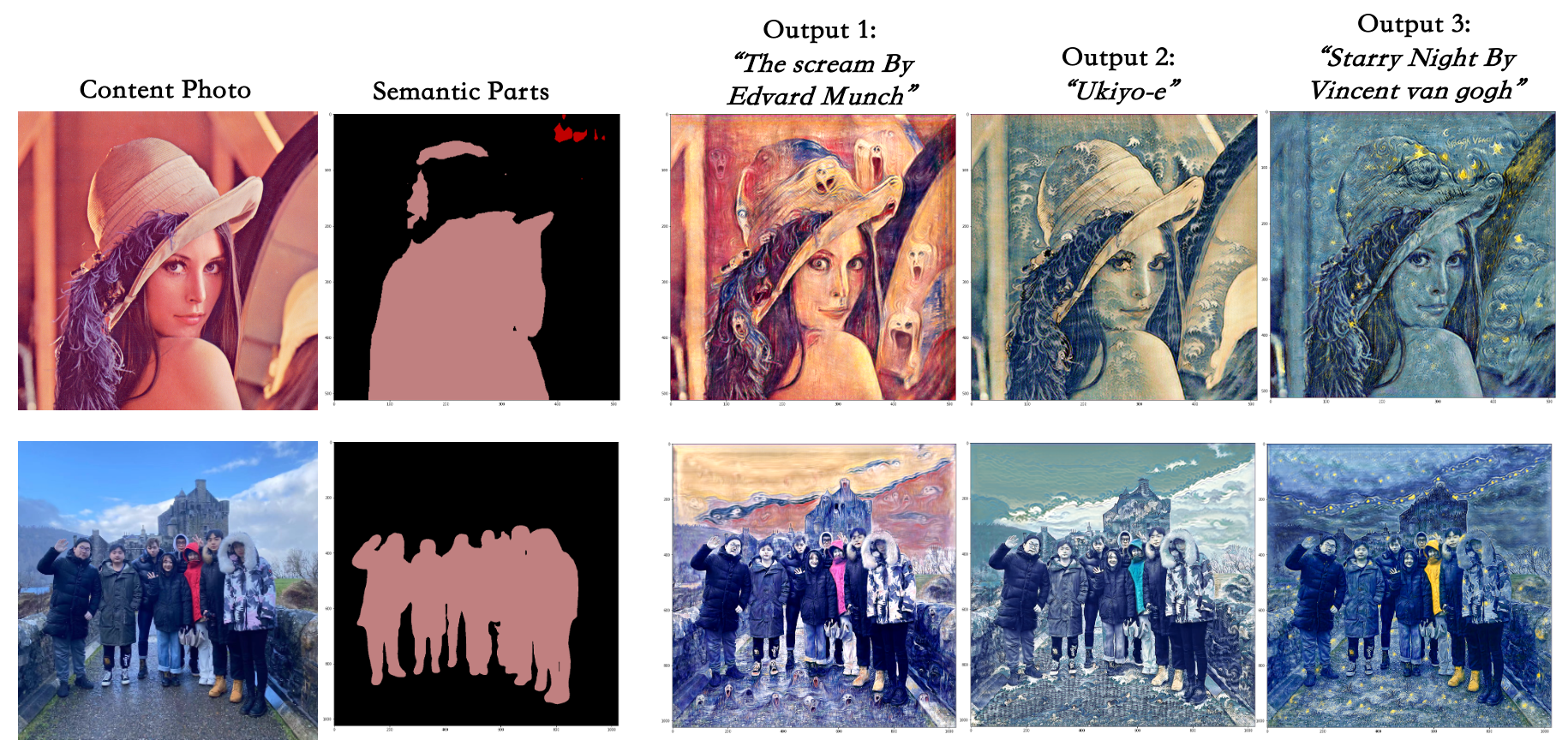}
 \end{center}
 \caption{Our style transfer results under various textual conditions: the generated images are created according to a semantic-aware mechanism and contain realistic texture effects and vibrant colors.}
 \end{figure}
}]

\begin{abstract} 
\textit{Recall that most of the current image style transfer methods require the user to give an image of a particular style and then extract that styling feature and texture to generate the style of an image, but there are still some problems: the user may not have a reference style image, or it may be difficult to summarise the desired style in mind with just one image. The recently proposed CLIPstyler has solved this problem, which is able to perform style transfer based only on the provided description of the style image. Although CLIPstyler can achieve good performance when landscapes or portraits appear alone, it can blur the people and lose the original semantics when people and landscapes coexist. Based on these issues, we demonstrate a novel framework that uses a pre-trained CLIP text-image embedding model and guides image style transfer through an FCN semantic segmentation network. Specifically, we solve the portrait over-styling problem for both selfies and real-world landscape with human subjects photos, enhance the contrast between the effect of style transfer in portrait and landscape, and make the degree of image style transfer in different semantic parts fully controllable. Our Generative Artisan resolve the failure case of CLIPstyler and yield both qualitative and quantitative methods to prove ours have much better results than CLIPstyler in both selfies and real-world landscape with human subjects photos. This improvement makes it possible to commercialize our framework for business scenarios such as retouching graphics software.}
\end{abstract} 

\section{Introduction}
\label{sec:intro}
The author Gatys et al. proposed a neural network-based style transfer algorithm in 2015 \cite{gatys_ecker_bethge_2015}, which was subsequently published on CVPR 2016 \cite{Gatys_2016_CVPR}. The algorithm caused quite a stir and quickly became the benchmark for all style transfer today. Although the image generated by the algorithm in this paper looks very pleasing to the user, it will be distorted when applied to the photo for style transfer, because the VGG network cannot preserve the content of the image well. 

In practical situations, the user may not have access to the reference style image for various reasons or cannot summarize the desired style of image. However, users are still interested in textures that 'mimic' the style image. For example, a user wishes to convert their photo into the style of a Monet or Van Gogh just by typing in a paragraph, without actually owning a painting by these famous painters. To overcome this limitation and create genuinely creative works of art, we should come up with a way to incorporate our imagination of style into a painting.

The limitations had been breached by CLIPstyler \cite{CLIPstyler}, which innovatively allows style transfer to be carried out 'without' a reference style image. The approach uses the recently proposed text-image embedding model - CLIP \cite{CLIP} to directly transfer semantic textures of the style description into images. A lightweight CNN network is trained to transform content images towards stylised images, allowing it to express textural information related to textual conditions and produce realistic and colourful results. Rather than optimising the style loss by using the feature maps generated by the network directly, the method optimizes the similarity between the CLIP model output of the images generated by the network and the textual conditions entered by the user. To reduce the content loss of the original image, data enhancement is applied by slicing the output image and sampling the patch with different perspective views. This enhancement allows for a more vivid and varied style of patches. Furthermore, to overcome the problem of patch-dependent over-stylization, a new threshold adjustment is applied so that patches with unusually high scores do not affect network training \cite{CLIPstyler}.

Although very high-quality images can be obtained with CLIPstyler, there are still many problems. For example, when stylizing people in a landscape picture, people are prone to being over-stylized and lose their original shape. The same is true for parts that contain much detail, such as folds of clothing, dense floor tiles, etc. These problems motivated our research, so we proposed specific research questions: to solve the problem of excessive stylization of portraits in selfies and real-world landscape with human subjects photos, to optimise the contrast of style transfer effect between portrait and landscape. In order to achieve our goals, we decided to use a semantic segmentation network to guide patchwise CLIP loss to get the optimal placement of patches and improve global CLIP loss by adding a portrait mask to enhance the contrast of style transfer effect between portrait and landscape. We propose a new semantic-aware and controllable patchwise CLIP loss for this purpose. Our experiments show that our new losses can achieve more reasonable and powerful style transfer than the original losses of CLIPstyler.

\section{Related Work}

\subsection{Neural Style Transfer}
Until now, neural style transfer methods for deep learning can be broadly classified into two categories: Image-Optimisation-Based Online Neural Methods (IOB-NST) and Model-Optimisation-Based Offline Neural Methods (MOB-NST). IOB-NST transfers the style by iteratively optimising an image; MOB-NST optimises the model and produces the stylised image with a single forward pass. 

\subsubsection{Image-Optimisation-Based Online Neural Methods (IOB-NST)}
The goal of IOB-NST is to make an empty white noise image match both content features in the content image and style features in the style image, resulting in a stylised content image. The method can generate high quality synthetic images without training the network. The parameters are also easy to adjust. However, it has a long computation time and great reliance on pre-trained models.

\textbf{Based on Maximum Mean Discrepancy:}
In 2015, Gatys et al. proposed a neural network-based style transfer algorithm \cite{Gatys15}, which was subsequently published on CVPR 2016 \cite{Gatys}. After applying a pre-trained VGG-19 network, Gatys et al. found that it is possible to extract content features from any arbitrary photograph by reconstructing the representations from intermediate layers. They also found that the network is capable of extracting style features by constructing the Gram matrix. The method optimises the image per-pixel by jointly minimising content and style losses. Subsequently, Li et al. \cite{LiWLH17} showed that matching the Gram matrices of feature maps is equivalent to minimise the Maximum Mean Discrepancy (MMD). 

\subsubsection{Model-Optimisation-Based Offline Neural Methods (MOB-NST)}
Although IOB-NST can generate high quality synthetic images, they suffer from computational inefficiency. In contrast, MOB-NST uses a large number of images to train generative models that can produce stylised images, which largely solves the problem of computational ineffiency in image style transfer. 

Inspired by Gatys et al., Johnson et al. \cite{JohnsonAL16} proposed to train a feed-forward network to solve the optimisation problem in real-time (i.e. real-time style transfer). Building on the Gatys et al. algorithm, Johnson et al. use perceptual loss functions to train a generative model for a particular style. In contrast to previous loss functions that used pixel-by-pixel comparisons when training generative models, the perceptual loss function squares the difference between high-level features extracted from the pre-trained VGG model. There are other efforts to achieve real-time style transfer which can transfer multiple or even arbitrary styles (\cite{LiFYWL017a}; \cite{abs-1802-06474}; \cite{abs-1808-04537} etc.). 

\section{Methodology}

\subsection{Basic Framework of CLIPstyler}
The CLIPstyler aims to transfer the semantic style of target text \(t_{sty}\) to the content image \(I_{c}\) through the pre-trained text-image embedding model CLIP \cite{CLIP}. Since the style is reflected in the form of text, we are able to perform image style transfer based only on the description of the style, without the need for a specific style image. 

\begin{figure}[h]
\centering
\includegraphics[width=\columnwidth]{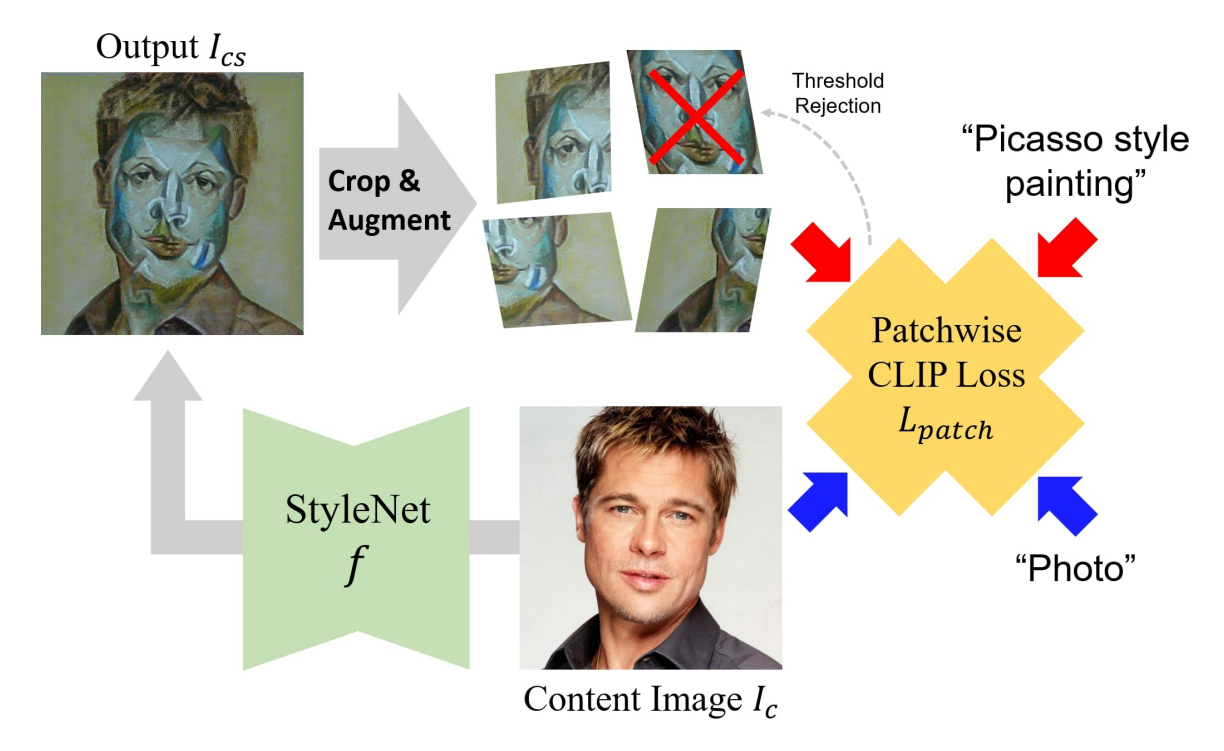}
\caption{CLIPstyler Architecture \cite{CLIPstyler}}
\label{CLIPstyler Architecture}
\end{figure}

The specified architecture of the network is shown in Figure \ref{CLIPstyler Architecture}. When a content image \(I_{c}\) is given, we aim to obtain the style transfer output \(I_{cs}\). In order to transfer the style to the content image, a CNN encoder-decoder model \(f\) (StyleNet) is used, which can capture the visual features of the content image and simultaneously stylise the image in deep feature space to obtain a realistic texture representation. Therefore, the stylised image \(I_{cs}\) is \(f(I_{c})\), and the final goal is to optimise the parameter of \(f\).

In 2016, \cite{Gatys} proposed to combine content loss and style loss function, which is still widely used nowadays in style transfer tasks. In order to optimise the neural network \(f\), content loss is used as a part of the loss function. However, since the style image is not provided, style loss cannot be computed. In order to achieve better performance, \citet{CLIPstyler} proposed a patchwise CLIP loss based on directional CLIP loss, combined these two losses with content loss and total variance regularisation loss, to form the following loss function: 
\begin{equation}
    L_{total} = \lambda_{d} L_{dir} + \lambda_{p} L_{patch} + \lambda_{c} L_{c} + \lambda_{tv} L_{tv}
\end{equation}
where $L_{total}$, $L_{dir}$, $L_{patch}$, $L_{c}$, $L_{tv}$ represents the total loss, directional CLIP loss, patchwise CLIP loss and total variation regularization loss, respectively. And $\lambda_{d}$, $\lambda_{p}$, $\lambda_{c}$, $\lambda_{tv}$ are weights for each of losses.

\subsection{Components of CLIPstyler}
\textbf{CLIP:} \cite{CLIP} CLIP is a state-of-the-art pretrained text-to-image embedding model. It is trained by predicting which caption goes with which image and enables the model to zero-shot transfer to downstream tasks. In the CLIPstyler, the CLIP pre-trained text transformer encoder and the vision transformer encoder are used to obtain both the CLIP space representation of image generated by StyleNet and the CLIP space representation of input text describing the style.

\begin{figure*}[ht]
\centering
\includegraphics[width=\linewidth]{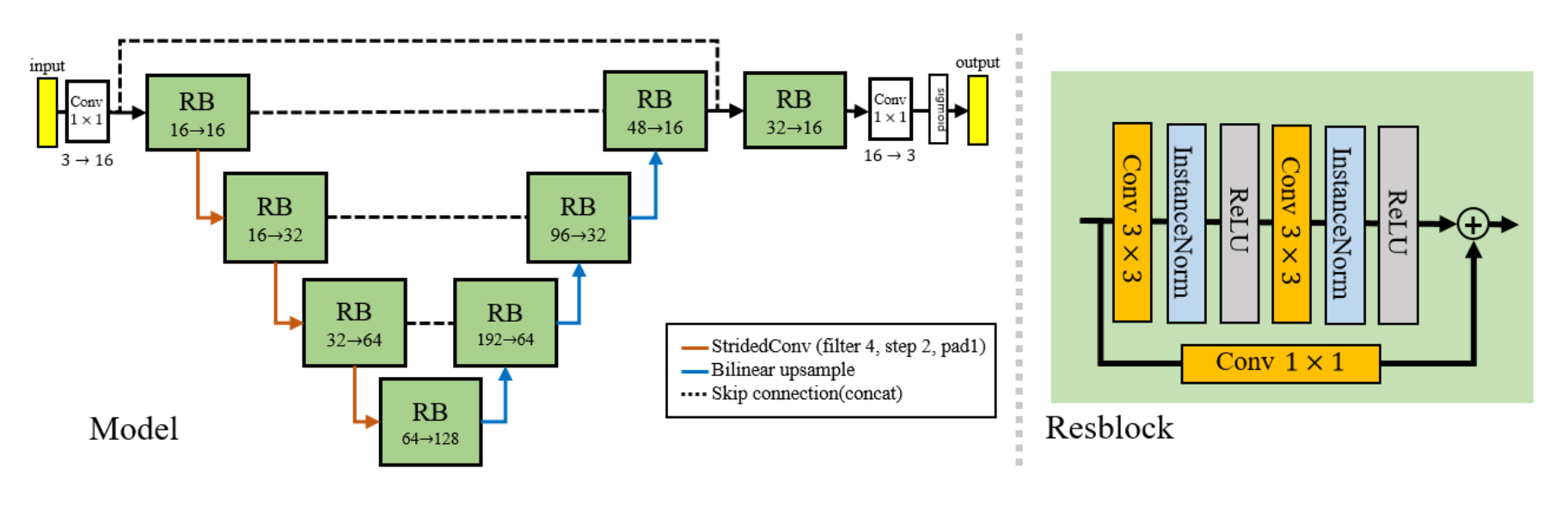}
\caption{StyleNet Model Architecture \cite{CLIPstyler}}
\label{StyleNet model architecture}
\end{figure*}

\textbf{StyleNet:} \cite{CLIPstyler} StyleNet is a CNN encoder-decoder model proposed by CLIPstyler. It captures hierarchical visual features of content images while stylizing images in a deep feature space for realistic texture representations. It uses a lightweight U-net structure and uses residual blocks to improve the content preservation and training stability. In the project, the StyleNet is trained to have the ability to transfer the semantic style of text to the content image. Model architecture is shown in Figure \ref{StyleNet model architecture}.

\textbf{Directional CLIP Loss:} \cite{CLIPstyler} Directional CLIP loss forces generated image to have a similar semantic style to the input text by aligning the CLIP-space direction between the text-image pairs of source and output. It optimizes the original CLIP loss to keep the style of image generated by StyleNet and semantic style of input text consistent with the original CLIP style. It can be defined by the following formula:
\begin{equation}
    \Delta T = E_{T}(t_{sty}) - E_{T}(t_{src})
\end{equation}
\begin{equation}
    \Delta I = E_{I}(f(I_{c})) - E_{I}(I_{c})
\end{equation}
\begin{equation}
    L_{dir} = 1 - \frac{\Delta I \cdot \Delta T}{\mid \Delta I \mid \mid \Delta T \mid}
\end{equation}
where $\Delta T$ represents the direction of CLIP space representation of input text and source text, $\Delta I$ represents the direction of CLIP space representation of stylized image and content image, and $L_{dir}$ represents the one minus cosine similarity between two directions mentioned above. 

\textbf{Patchwise CLIP Loss:} \cite{CLIPstyler} Patchwise CLIP loss uses randomly cropped patches instead of whole generated image to compute directional CLIP loss. It calculates the directional CLIP loss separately after prescriptive augmentation for each randomly cropped patch, and calculates the average of the loss of all patches after rejecting overly stylized patches (threshold regularization). It can be defined by the following formula:
\begin{equation}
    \Delta T = E_{T}(t_{sty}) - E_{T}(t_{src})
\end{equation}
\begin{equation}
    \Delta I = E_{I}(aug(\hat{I}_{cs}^{(i)})) - E_{I}(I_{c})
\end{equation}
\begin{equation}
    l_{patch}^{(i)} = 1 - \frac{\Delta I \cdot \Delta T}{\mid \Delta I \mid \mid \Delta T \mid}
\end{equation}
\begin{equation}
    L_{patch} = \frac{1}{N} \sum_{i}^{N} R(l_{patch}^{(i)},\tau)
\end{equation}
\begin{equation}
    \text{where } R(s,\tau) = \begin{cases}
                0 & \text{if s $<= \tau$} \\
                s & \text{otherwise}
               \end{cases}
\end{equation}
where $\Delta T$ represents the direction of CLIP space representation of input text and source text, $\Delta I$ represents the direction of CLIP space representation of a patch from stylized image after augmentation and content image, and $l_{patch}^{(i)}$ represents the one minus cosine similarity between two directions mentioned above. $L_{patch}$ represents the average of all patch losses after doing threshold rejection with threshold value $\tau$. The patches with loss lower than $\tau$ will be nullified.

\textbf{Content Loss:} \cite{Gatys} Content loss computes the mean squared error between content image and generated image on content features extracted from a pretrained VGG-19 network, which was firstly proposed by Gatys et al. In the CLIPstyler, content loss is obtained by adding the mean squared error of the content features of Conv\_4\_2 and Conv\_5\_2 layers, which is used to preserve the content of the original image on generated image.

\textbf{Total Variance Regularization Loss:} \cite{regular} Total variation regularization loss is inspired by noise removal technique. It uses mean squared error to minimize the difference between adjacent pixels, thus alleviate the side artifacts from irregular pixels.

\textbf{Multiview Augmentation:} \cite{perspective} Multiview augmentation performs a random perspective transformation on a given image with a given probability. In the CLIPstyler, RandomPerspective implemented by Pytorch torchvision library is used with degree of distortion 0.5 in all experiments.

\textbf{Threshold Regularization:} \cite{CLIPstyler} Threshold regularization simply invalidates patches whose directional CLIP loss is below a certain threshold value during gradient optimization process, thus avoiding generated images suffering from over-stylization. In the CLIPstyler, threshold value is set to 0.7 in all experiments.

\textbf{Prompt Engineering:} \cite{CLIP} Prompt engineering technique combines some texts of similar meaning with the target text and feeds them to the text encoder, and then uses the averaged embedding to replace the original embedding with single text condition, thereby reducing the noise of the text embedding. In the CLIPstyler, 79 similar text templates are used to compute the averaged embedding. 

\subsection{Our Proposed Methods}

\begin{figure}[h]
\centering
\includegraphics[width=\columnwidth]{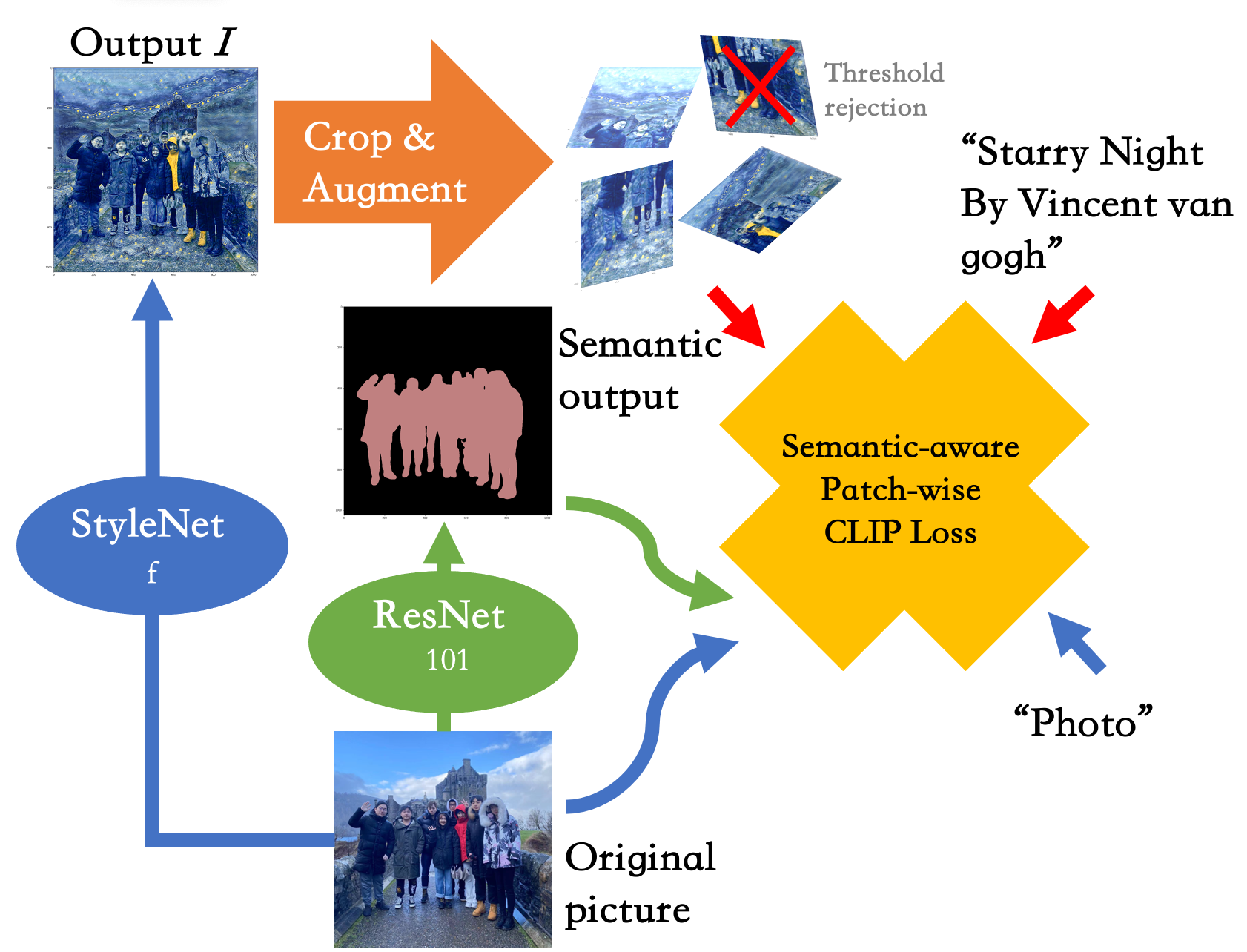}
\caption{Our New CLIPstyler Architecture}
\label{Our New CLIPstyler Architecture}
\end{figure}

During experiments, we found that the CLIPstyler will over-stylize portraits when portraits and landscapes coexist, especially the distortion of human faces (Figure \ref{A comparison of the generated results of our methods - Generative Artisan CLIPstyler with the original version - Base CLIPstyler in the absence of the (background) global CLIP loss.}). We believe this is caused by the fact that patchwise CLIP loss randomly stylizes different regions, resulting in some undesired regions being overly stylized. In addition, due to the large difference between landscape and portrait, global CLIP loss cannot achieve the desired effect on images with both portrait and landscape. To address these issues, we propose to use a semantic segmentation network to guide the style transfer of CLIPstyler, which makes style transfer semantically aware and controllable. Our new CLIPstyler architecture is shown in Figure \ref{Our New CLIPstyler Architecture}.

\begin{figure}[h]
\centering
\includegraphics[width=\columnwidth]{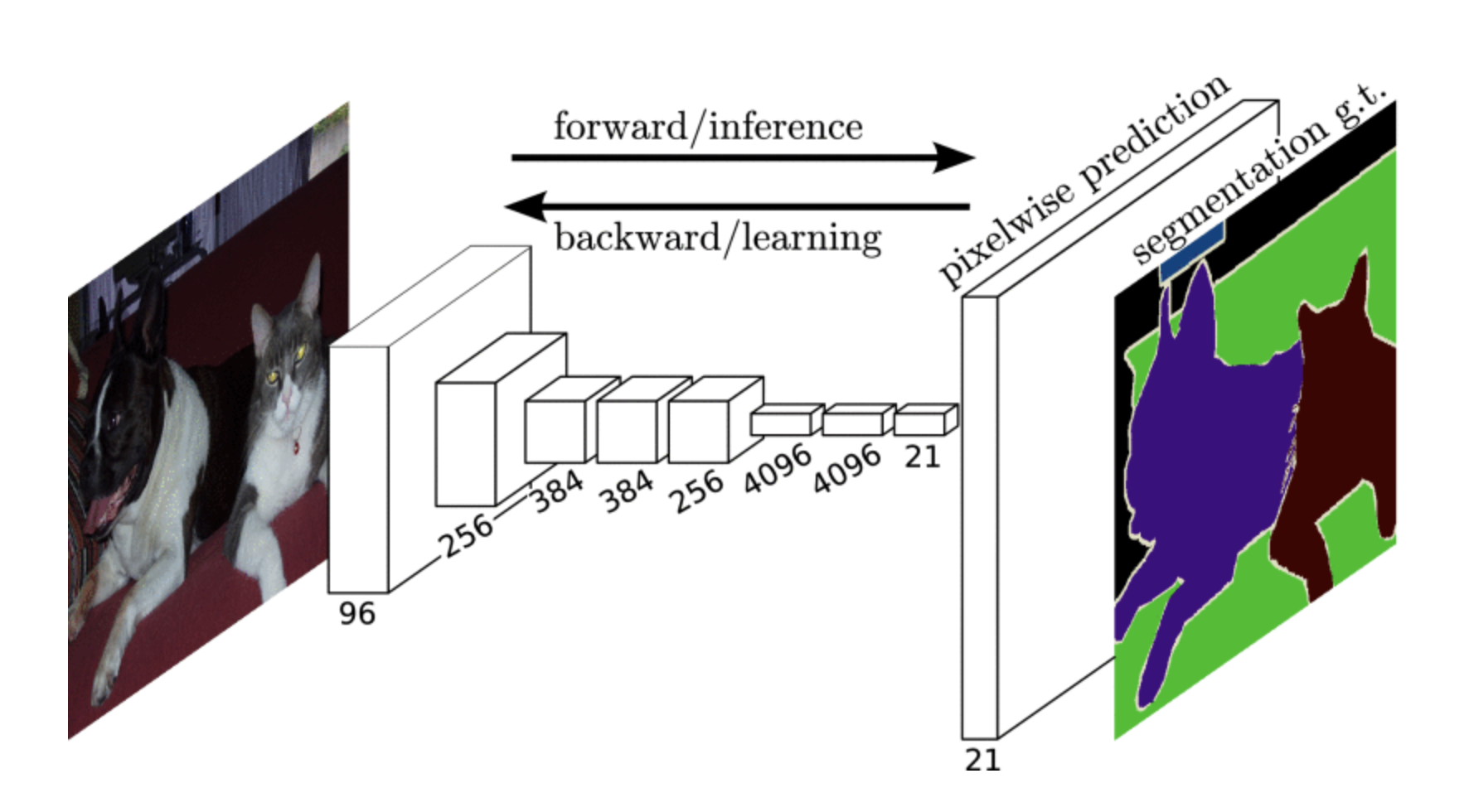}
\caption{Semantic Segmentation Network based on Fully-Convolutional Network \cite{FCN}}
\label{fig:FCN}
\end{figure}

\textbf{Semantic Segmentation Network:} In order to solve the problem that CLIPstyler cannot make portraits less stylized while making background more stylized. We propose a method that performs semantic segmentation of portrait and background using the semantic segmentation network (Figure \ref{fig:FCN}) proposed by Evan Shelhamer et al. \cite{FCN}. They proposed to use a Resnet101 as the backbone for end-to-end semantic segmentation. The reason for choosing this network is that it performs well on semantic segmentation of human portraits. This network will provide information on the location of different things in the image for our refinement.

\textbf{Semantic-aware Patchwise CLIP Loss:} We propose a semantic-aware patchwise CLIP loss, which applies a pre-trained FCN network to obtain semantic information of portraits and backgrounds in the original image, and uses them to determine the degree of style transfer for different semantic parts. We propose to use the following mechanisms to control the degree of style transfer across different semantic parts:
\begin{itemize}
    \item \textbf{Semantic-aware Random Cropping:} In the original patchwise CLIP loss, random cropping does not have any restrictions, so it is very easy to crop to areas that are not suitable for style transfer, resulting in corrupted images. We propose to use the semantic information in the image to control the number of patches for different semantic parts. We have experimentally verified that using fewer patches for portraits and more patches for landscapes can increase the degree of style transfer for landscapes while avoiding damage to portraits.
    \item \textbf{Semantic-aware Weight Penalty:} In the original patchwise CLIP loss, every patch that is not rejected is treated equally. In this way, different degrees of style transfer cannot be performed according to different semantic parts. Therefore we propose to use different weight penalties for patches in different semantic parts. We experimentally verify that using a higher weight penalty for the patch in portrait part and a lower weight penalty for the patch in landscape part can make the style transfer of the landscape better without destroying the portrait.
    \item \textbf{Semantic-aware Threshold Rejection:} In the original patchwise CLIP loss, threshold rejection is a fixed value for each patch. This means that it does not take into account that different semantic parts should redefine more appropriate thresholds. We experimentally verify that using a higher threshold value for the patch in portrait part and a lower threshold value for the patch in landscape part can make the style transfer of the landscape better without destroying the portrait.
    \item \textbf{Semantic-aware Regulatable Patch:} We propose a regulatable patch that can determine which semantic part a patch belongs to based on only part of the patch region. This method is proposed because the original patch size of patchwise CLIP loss is fixed, which will lead to obvious differences at the junction of different semantic parts. In order to maintain the uniformity of the picture style, we can only use a small part of the patch to decide that the patch belongs to the portrait area. We experimentally verify that such a setting can make the effect of image style transfer more uniform.
\end{itemize}
The definition of Semantic-aware Patchwise CLIP Loss can be formulated as follow:
\begin{equation}
    \Delta T = E_{T}(t_{sty}) - E_{T}(t_{src})
\end{equation}
\begin{equation}
    \Delta I = E_{I}(aug(\hat{I}_{cs}^{(i)})) - E_{I}(I_{c})
\end{equation}
\begin{equation}
    l_{patch}^{(i)} = 1 - \frac{\Delta I \cdot \Delta T}{\mid \Delta I \mid \mid \Delta T \mid}
\end{equation}
\begin{equation}
    L_{patch} = \frac{1}{N} \sum_{i}^{N} {W(l_{patch}^{(i)},\alpha)} * R(l_{patch}^{(i)},\tau)
\end{equation}
\begin{equation}
    \text{where } R(s,\tau) = \begin{cases}
                0 & \text{if s $<= \tau$} \\
                s & \text{otherwise}
               \end{cases}
\end{equation}
\begin{equation}
    \text{where } \tau = \begin{cases}
                \tau_{portrait} & \text{if patch belongs to portrait} \\
                \tau_{back} & \text{if patch belongs to back}
                \end{cases}
\end{equation}
\begin{equation}
    \text{where } W(s,\alpha) = \begin{cases}
                \alpha_{portrait} & \text{if $s$ belongs to portrait} \\
                \alpha_{back} & \text{if $s$ belongs to back}
               \end{cases}
\end{equation}
\begin{equation}
    \text{where the number of } \hat{I}_{cs}^{(i)} \text{ in portrait is constrained}
\end{equation}
\begin{equation}
    \text{where size of } \hat{I}_{cs}^{(i)} \text{ to determine part is constrained}
\end{equation}

The semantic-aware patchwise CLIP loss solves the problem that patchwise CLIP loss in the CLIPstyler randomly crop patches, resulting in bad patches that distorts undesired regions such as faces. Using the semantic segmentation network to guide the degree of style transfer of different patches can effectively avoid the influence of bad patches at locations where excessive style transfer is not expected. Not only that, but it also means that different semantically segmented parts can change their degree of style transfer according to our artificial settings, making the image style transfer freely controllable.

\textbf{Background Global CLIP Loss:} In order to enable global CLIP loss in the CLIPstyler to pay more attention to the style transfer of the background without damaging the portrait, we propose to use a mask for the portrait part of the generated image, so that the pixels of the portrait part are not affected by the global CLIP loss. At the same time, we use the generated image containing the portrait mask to replace the original generated image as the input of CLIP, and only make the style of the background part match the semantic style of the input text. The Background Global CLIP Loss is formulated as follow: 
\begin{equation}
    \Delta T = E_{T}(t_{sty}) - E_{T}(t_{src})
\end{equation}
\begin{equation}
    \Delta I = E_{I}(Mask(f(I_{c}))) - E_{I}(Mask(I_{c}))
\end{equation}
\begin{equation}
    L_{dir} = 1 - \frac{\Delta I \cdot \Delta T}{\mid \Delta I \mid \mid \Delta T \mid}
\end{equation}

where $Mask$ represents portrait mask which can be defined using the following formula:
\begin{equation}
    Mask = \begin{cases}
       0 & \text{if pixel belong to the portrait} \\
       1 & \text{if pixel belong to the back}
       \end{cases} 
\end{equation}
Our experiments found that this approach not only effectively avoids damage to the portrait, but also makes the style transfer effect of the background better than that of CLIPstyler. This may be because it is easier for CLIPstyler to match the text semantic style with the background style when only the background is present.

The above two losses enable CLIPstyler to achieve results in pictures greatly improving the original effect of CLIPstyler. Based on our methods, the loss function for training CLIPstyler is modified as follows:
\begin{equation}
    L_{total} = \lambda_{d} L_{back\_dir} + \lambda_{p} L_{sem\_patch} + \lambda_{c} L_{c} + \lambda_{tv} L_{tv}
\end{equation}
where $L_{back\_dir}$ represents background global CLIP loss and $L_{sem\_patch}$ represents semantic-aware patchwise CLIP loss.

\section{Data Set and Evaluation Criteria} 
\subsection{Data}
Since our model is based on image optimisation (IOB-NST), the model only needs one image as input and does not need to use training set, validation set and test set. In this project, we will use the template portrait pictures provided by CLIPstyler and real real-world landscape with human subjects photos (e.g. sky + mountains + people + ground, etc.) to test the performance of the model.

\subsection{Evaluation}
\subsubsection{Quantitative Evaluation}
In CLIPstyler, there is no quantitative evaluation method provided for model performance directly. We decided to evaluate the performance of style transfer by using four different losses provided in CLIPstyler, as well as our newly proposed two losses. In theory, lower values of these losses would represent better style transfer. For example, the two CLIP losses of CLIPstyler are used to measure the effect of style transfer, the content loss is used to measure the degree to which the original image content is preserved, and the total variation regularization loss is used to measure the impact of irregular pixels. If all the losses of our optimized image are lower than the baseline image, it will mean that our methods will be completely better than the baseline.

\subsubsection{Qualitative Evaluation}
We used both visual comparison and ablation experiments for qualitative evaluation. We decided to use several well-known painter style descriptions such as \emph{'The Scream by Edvard Munch'} to evaluate the difference between our optimized images and baseline images. During the experiment, we found that the style transfer effect of the baseline is extremely unstable, which may be caused by the ambiguity of the style description. Therefore, we choose the picture with the best quality generated by the baseline as the comparison object. We also select optimized images similar to the baseline images for comparison. 

For ablation experiments, we found that there is no difference to the original image without using any CLIP loss or only using the global CLIP loss, so we decided to investigate whether our optimized global CLIP loss can make the style transfer effect of background stronger by only discarding the global CLIP loss, and whether our optimized global CLIP loss plays a more important role in the style transfer process than baseline ones. Furthermore, we can also compare the performance difference between our proposed patchwise CLIP loss and the original patchwise CLIP loss.

\section{Experiments}

\begin{figure*}[h]
\begin{center}
\includegraphics[width=\linewidth]{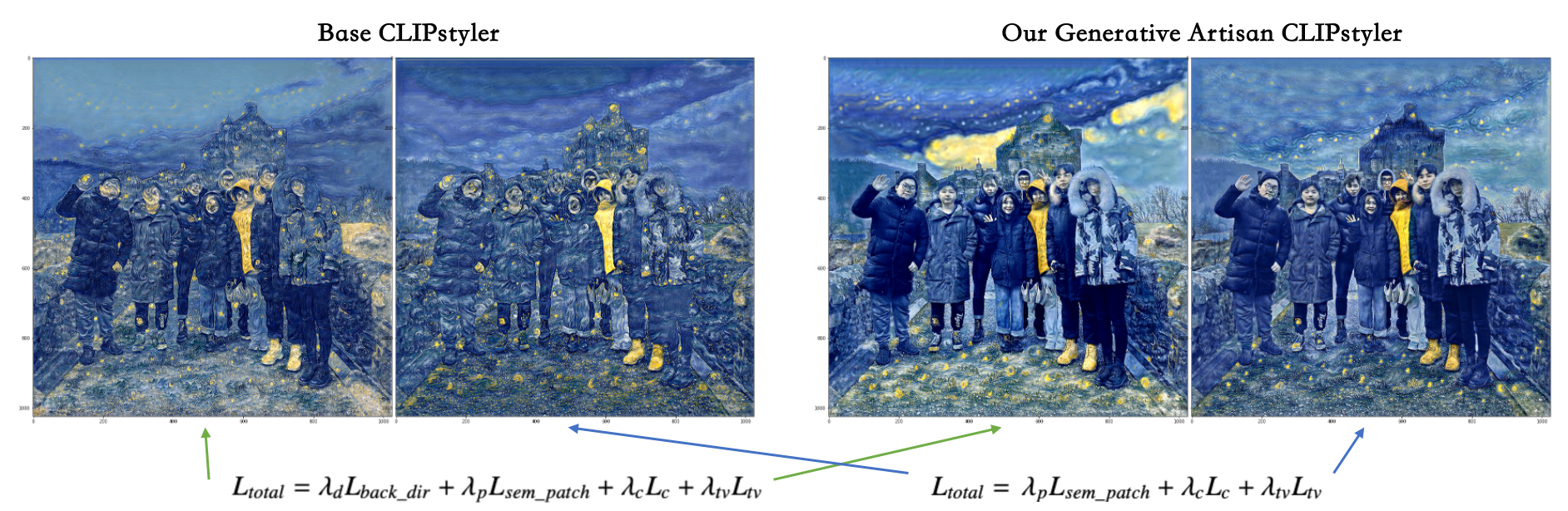}
\end{center}
\caption{A comparison of the generated results of our methods - Generative Artisan CLIPstyler with the original version - Base CLIPstyler in the absence of the (background) global CLIP loss.}
\label{A comparison of the generated results of our methods - Generative Artisan CLIPstyler with the original version - Base CLIPstyler in the absence of the (background) global CLIP loss.}
\end{figure*}

\subsection{Experimental setup}
We use 512x512 resolution for all template images provided by CLIPstyler and 1024x1024 HD images for real photos. For training, we set $\lambda d$, $\lambda p$, $\lambda c$, $\lambda v$ to be 5 × $10^{2}$, 9 × $10^{3}$, 150, and 2 × $10^{-3}$, respectively. For content loss, we use features from conv4\_2 and conv5\_2 layers to handle content loss. We use the lightweight U-net architecture provided by CLIPstyler as StyleNet, and use the Adam optimizer with a learning rate of 5 × $10^{-4}$ to train the model. The model iteratively optimize the pixels of the input image in each epoch and our total number of training iterations is set to 400 epochs and the learning rate is reduced to half every 100 iterations. We train the model on a single P100 GPU, and the training time for each style transfer is about 5 minutes. For patch cropping, we use a patch size of 128 as our default setting, and the total number of cropped patches is set to n = 64. For perspective argumentation, we use functions provided by the Pytorch library. For threshold rejection, we set threshold value to 0.7 in baseline model, while for optimized loss, the size of threshold will be parameterized according to the specific photo. To reduce the noise of text embeddings, we use prompt engineering techniques mentioned in methodology. Finally, to present better visual effects to readers, we apply the contrast enhancement technique which enhances image contrast 1.5 times to all outputs including the baseline.

\subsection{Experimental results}

\subsubsection{Comparison with Baseline}
In a qualitative way, we decide to visually compare the images generated by our model with those generated by the base CLIPstyler, which is shown in Appendix Figure \ref{A comparison of the generated results of our methods with the original version. The first and third rows are by Basic CLIPstyler, and the second and fourth rows are by Generative Artisan CLIPstyler.}. In general, we have perfectly solved the problem of the style having over-specified representations on the portrait. 

In the first column of Figure \ref{A comparison of the generated results of our methods with the original version. The first and third rows are by Basic CLIPstyler, and the second and fourth rows are by Generative Artisan CLIPstyler.}, we have applied the style in \emph{The Scream} by Edvard Munch to our real-world landscape with human subjects photo, and Lena, respectively. As we can see, the images generated by the base CLIPstyler have over-specified the element of "scream" and apply it on human faces, which causes a very serious distortion. In contrast, our model removes the effect of the scream expression on the portraits, while making the background section more stylised. In our real-world landscape with human subjects photo, we can see that more elements of scream appear in the sky and on the ground, making the overall image look more "screamy". This optimisation is also evident in the second column (the style of \emph{The Starry Night} by Vincent van Gogh) and the third column (the style of \emph{The Great Wave off Kanagawa} by Katsushika Hokusai) of the generated images. In the top two images in the second column, the image generated by the base CLIPstyler has a lot of stars covering the portrait, while few stars are visible in the background sky. After optimisation, it can be seen that the human portraits are clearly shown and the sky becomes closer to that of the Van Gogh's work. A similar optimisation is seen in the two images below in the second column, where the swirling texture on the face disappears and is replaced by a more complete portrait. 

\subsubsection{Ablation Experiment Results}
The results of our ablation experiments are shown in Figure \ref{A comparison of the generated results of our methods - Generative Artisan CLIPstyler with the original version - Base CLIPstyler in the absence of the (background) global CLIP loss.}. The figure shows the images of the starry night style when using and not using the global CLIP loss. The image pair on the left are generated from base CLIPstyler and the image pair on the right are generated from Generative Artisan CLIPstyler. For each pair of images, the image on the left is with global CLIP loss, and the image on the right is without global CLIP loss. For the image pair of Generative Artisan CLIPstyler, we use our own proposed CLIP loss instead of the original CLIP loss to compare with the baseline.

From the image pair of base CLIPstyler, we can find that there is not much difference between images with and without using global CLIP loss. This verifies that the global CLIP loss mentioned in CLIPstyler cannot achieve the desired effect. However, in the image pair of Generative Artisan CLIPstyler, we can find that our proposed background global CLIP loss can greatly improve the degree of style transfer of the background. This verifies that using a portrait mask can effectively improve the degree of style transfer of the background by making the text semantics match with the image style of background only. This also means that the background global CLIP loss can play a more important role in style transfer than the original global CLIP loss.

Comparing the images of base CLIPstyler and Generative Artisan CLIPstyler using only patchwise CLIP loss, we can find that the image of base CLIPstyler obviously damages the portrait, which verifies that the randomly cropped patch is indeed easy to transfer excessive texture to the human face. However, the image of Generative Artisan CLIPstyler perfectly preserves the portrait while optimizing the contrast between the portrait and the background. This verifies that the semantic-aware patchwise CLIP loss can avoid excessive style transfer of portraits and improve the contrast of the degree of style transfer between portrait and background.

Overall, our newly proposed CLIP loss outperforms the original CLIP loss in all aspects of visual performance. Not only does this allow our portraits to retain a similar quality to a real photo, but it also enriches the stylistic elements of the landscape.

\subsubsection{Quantitative Experimental Results}
In addition to judging which style-transferred image is better by visual comparison, we also need to judge the pros and cons by model performance. The figures below shows the different loss variations for two different image examples and the comparison between the individual losses.

\begin{figure}[h]
\begin{center}
\includegraphics[width=\linewidth]{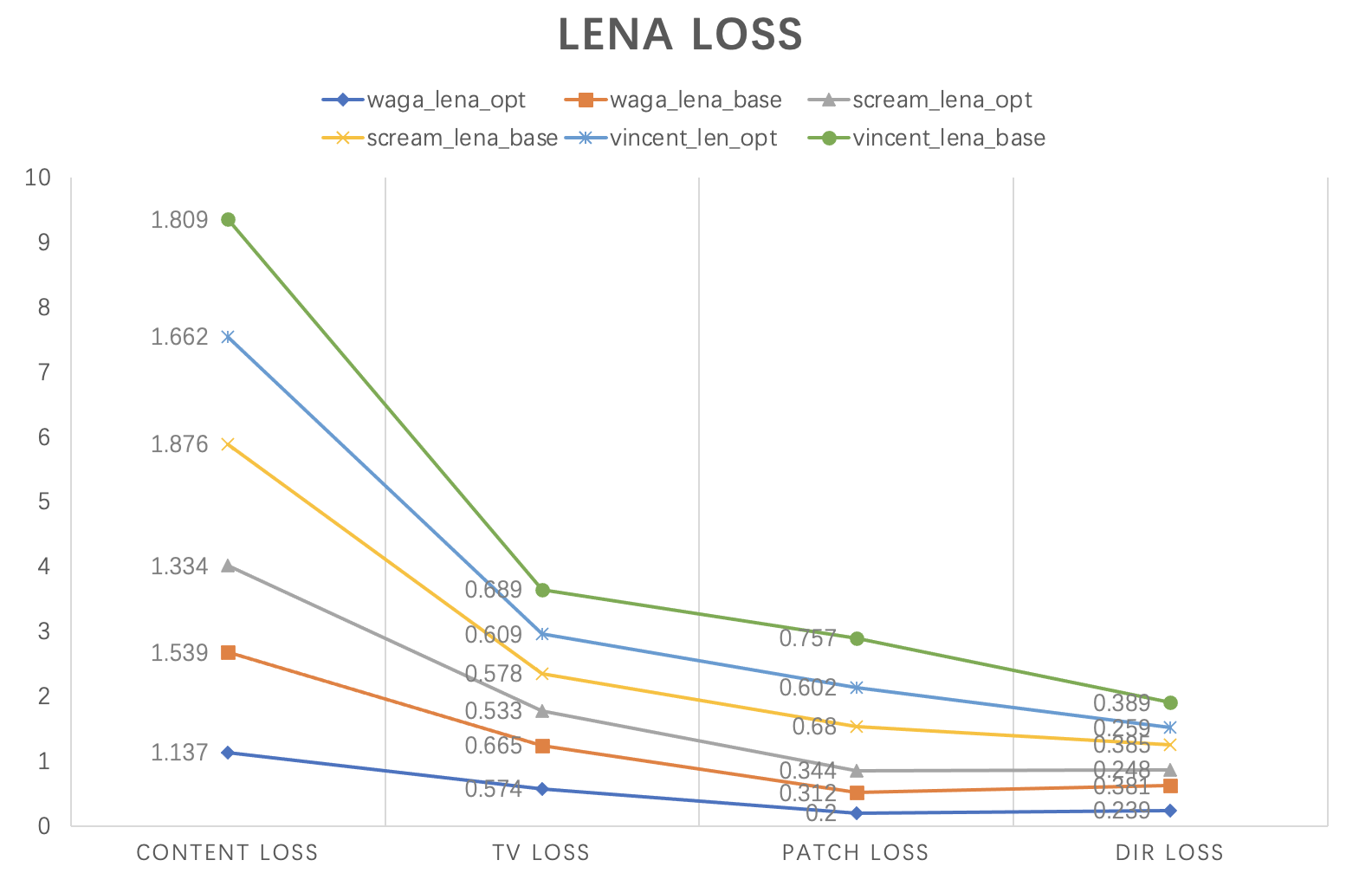}
\end{center}
\caption{Lena different loss line graph.}
\label{Lena different loss line graph.}
\end{figure}
Figure \ref{Lena different loss line graph.} shows the results of Lena picture we used for our experiments. We can see that no matter what type of style transfer, all the losses of the optimized images are lower than the baseline ones. This shows that the style transfer effect of our model on Lena picture is better than the baseline style transfer effect in all aspects. 

\begin{figure}[h]
\begin{center}
\includegraphics[width=\linewidth]{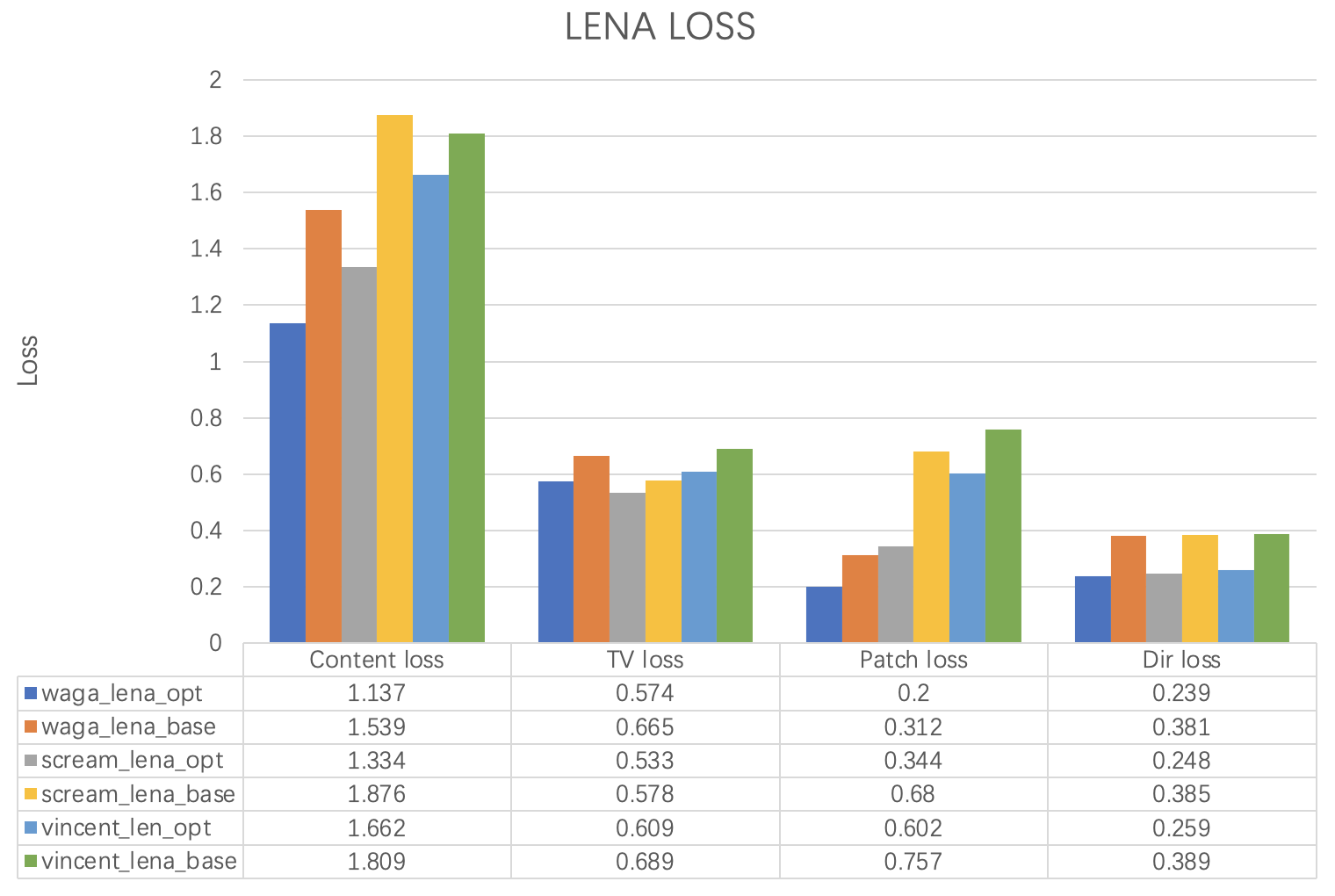}
\end{center}
\caption{Lena different loss bar graph.}
\label{Lena different loss bar graph.}
\end{figure}
Figure \ref{Lena different loss bar graph.} shows the comparison of all individual losses in Lena. we can find that the optimised content loss is much lower compared to baseline and the optimized CLIP losses are slightly lower than the baseline. This meets our expectations, as our goal is to have better style transfer than the original CLIPstyler while preserving the full portrait. The reduction in TV loss also shows that there are fewer irregular pixels left in the image.

\begin{figure}[h]
\begin{center}
\includegraphics[width=\linewidth]{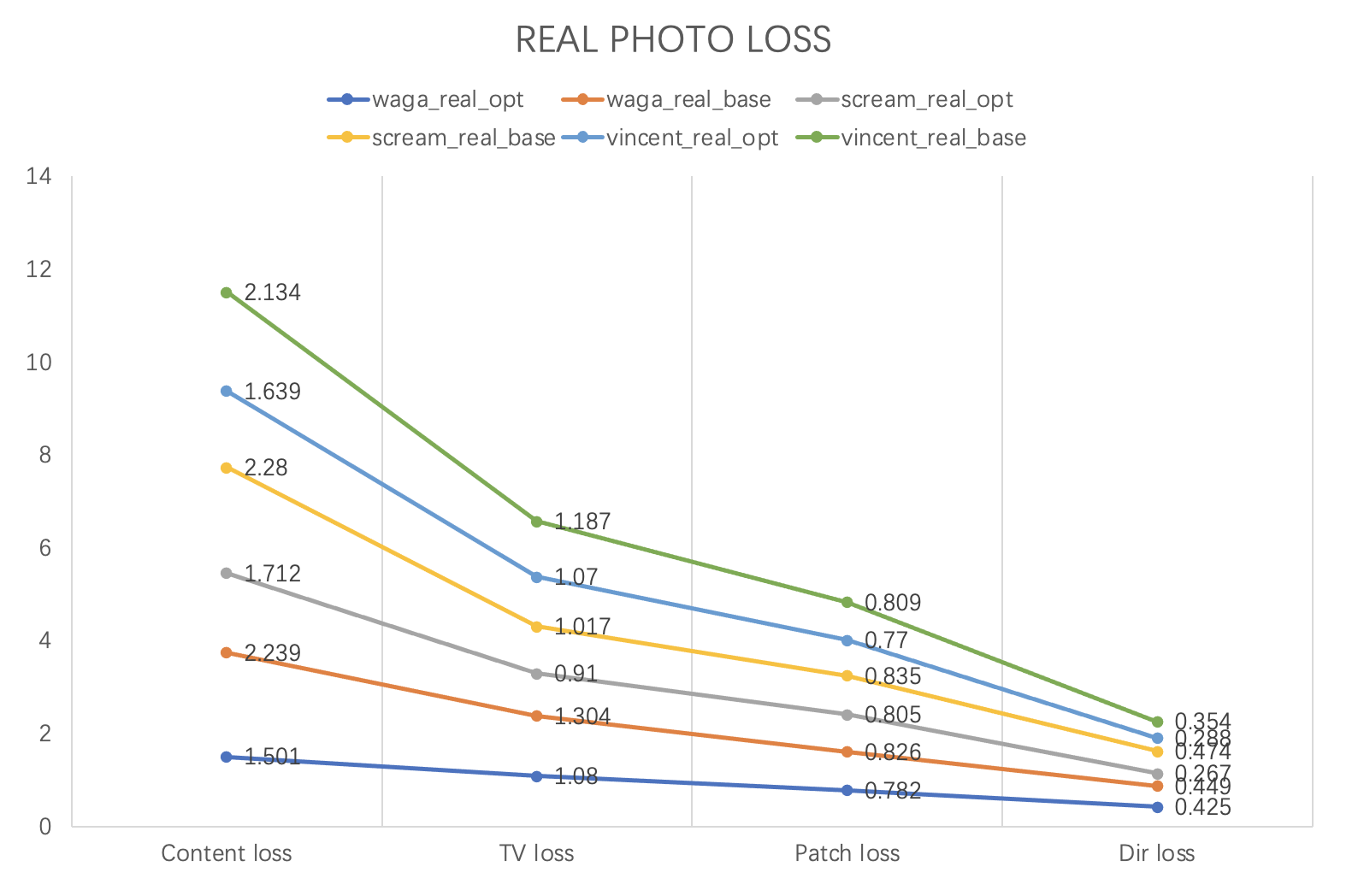}
\end{center}
\caption{real-world landscape with human subjects photo different loss line graph.}
\label{real-world landscape with human subjects photo different loss line graph.}
\end{figure}
Figure \ref{real-world landscape with human subjects photo different loss line graph.} shows the results of real-world landscape with human subjects photos we used for our experiments. We can also see that no matter what type of style transfer, all the losses of the optimized images are lower than the baseline ones. This shows that the style transfer effect of our model on real-world landscape with human subjects photo is also better than the baseline style transfer effect in all aspects.

\begin{figure}[h]
\begin{center}
\includegraphics[width=\linewidth]{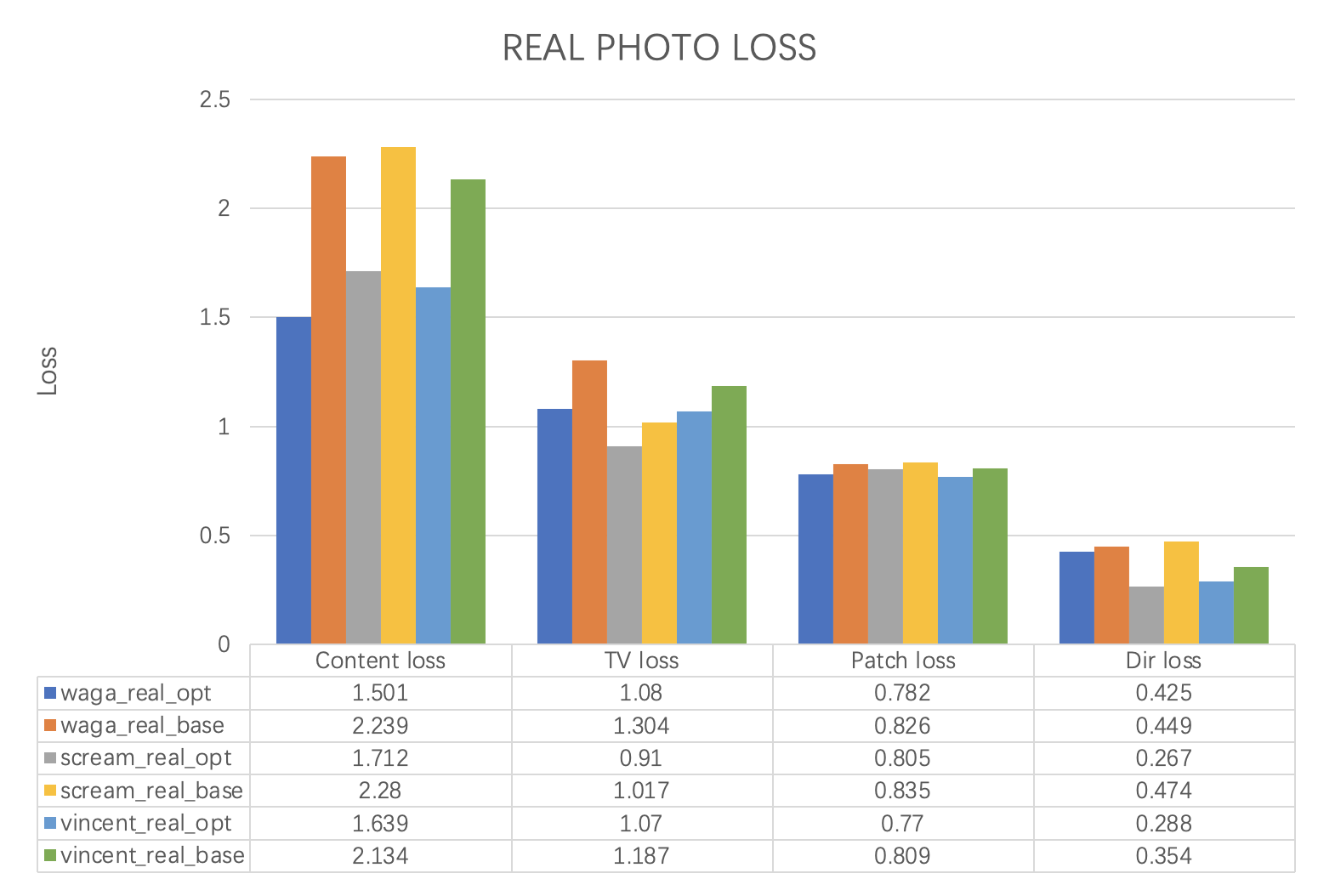}
\end{center}
\caption{real-world landscape with human subjects photo different loss bar graph.}
\label{real-world landscape with human subjects photo different loss bar graph.}
\end{figure}
Figure \ref{real-world landscape with human subjects photo different loss bar graph.} shows the comparison of all individual losses in real-world landscape with human subjects photo. We can get almost the same analysis results as the Lena, i.e. all losses outperform the baseline. This shows that the generalization performance of our model is also reflected in the real-world landscape with human subjects photo, which in a quantitative comparison proves that our model works optimally.

\section{Conclusions}
In this paper, we proposed a novel image style transfer framework that allows the user to input a piece of text to change the current image style without a reference style image, and incorporate a semantic segmentation network to guide the style transfer so that it focuses on the style transfer effects of portraits and landscapes, respectively. We proposed two losses based on semantic segmentation networks, namely the semantic-aware patchwise CLIP loss and the background global CLIP loss. These two losses addresses major issues with CLIPstyler, such as pathwise CLIP loss being too random and prone to excessive style transfer for undesired regions, as well as addressing some failure case in CLIPstyler and obtained much better results in some scenarios such as selfies and real-world landscape with human subjects photos. This improvement makes the image style transfer of CLIPstyler full controllable based on different semantic parts and makes CLIPstyler commercially viable for more business scenarios such as retouching software. Although we have the excellent results, the current technology is limited to backgrounds and portraits. There is no control over style transfer for more fine-grained background elements. In future work, we can use a more advanced semantic segmentation network to obtain richer semantic information, so that we can control more semantic parts for different degrees of style transfer, so as to obtain more brilliant results. In addition, we could also measure image quality via non-reference metrics e.g. FID or NIQE scores.

\bibliography{main}

\begin{thebibliography}{14}
\providecommand{\natexlab}[1]{#1}
\providecommand{\url}[1]{\texttt{#1}}
\expandafter\ifx\csname urlstyle\endcsname\relax
  \providecommand{\doi}[1]{doi: #1}\else
  \providecommand{\doi}{doi: \begingroup \urlstyle{rm}\Url}\fi

\bibitem[Gatys et~al.(2015{\natexlab{a}})Gatys, Ecker, and Bethge]{Gatys15}
Gatys, Leon~A., Ecker, Alexander~S., and Bethge, Matthias.
\newblock A neural algorithm of artistic style.
\newblock \emph{CoRR}, abs/1508.06576, 2015{\natexlab{a}}.
\newblock URL \url{http://arxiv.org/abs/1508.06576}.

\bibitem[Gatys et~al.(2015{\natexlab{b}})Gatys, Ecker, and
  Bethge]{gatys_ecker_bethge_2015}
Gatys, Leon~A., Ecker, Alexander~S., and Bethge, Matthias.
\newblock A neural algorithm of artistic style, Sep 2015{\natexlab{b}}.
\newblock URL \url{https://arxiv.org/abs/1508.06576}.

\bibitem[Gatys et~al.(2016{\natexlab{a}})Gatys, Ecker, and Bethge]{Gatys}
Gatys, Leon~A, Ecker, Alexander~S, and Bethge, Matthias.
\newblock Image style transfer using convolutional neural networks.
\newblock In \emph{Proceedings of the IEEE conference on computer vision and
  pattern recognition}, pp.\  2414--2423, 2016{\natexlab{a}}.

\bibitem[Gatys et~al.(2016{\natexlab{b}})Gatys, Ecker, and
  Bethge]{Gatys_2016_CVPR}
Gatys, Leon~A., Ecker, Alexander~S., and Bethge, Matthias.
\newblock Image style transfer using convolutional neural networks.
\newblock In \emph{Proceedings of the IEEE Conference on Computer Vision and
  Pattern Recognition (CVPR)}, June 2016{\natexlab{b}}.

\bibitem[Johnson et~al.(2016)Johnson, Alahi, and Fei{-}Fei]{JohnsonAL16}
Johnson, Justin, Alahi, Alexandre, and Fei{-}Fei, Li.
\newblock Perceptual losses for real-time style transfer and super-resolution.
\newblock \emph{CoRR}, abs/1603.08155, 2016.
\newblock URL \url{http://arxiv.org/abs/1603.08155}.

\bibitem[Kwon \& Ye(2021)Kwon and Ye]{CLIPstyler}
Kwon, Gihyun and Ye, Jong~Chul.
\newblock Clipstyler: Image style transfer with a single text condition.
\newblock \emph{arXiv preprint arXiv:2112.00374}, 2021.

\bibitem[Li et~al.(2018{\natexlab{a}})Li, Liu, Kautz, and Yang]{abs-1808-04537}
Li, Xueting, Liu, Sifei, Kautz, Jan, and Yang, Ming{-}Hsuan.
\newblock Learning linear transformations for fast arbitrary style transfer.
\newblock \emph{CoRR}, abs/1808.04537, 2018{\natexlab{a}}.
\newblock URL \url{http://arxiv.org/abs/1808.04537}.

\bibitem[Li et~al.(2017{\natexlab{a}})Li, Wang, Liu, and Hou]{LiWLH17}
Li, Yanghao, Wang, Naiyan, Liu, Jiaying, and Hou, Xiaodi.
\newblock Demystifying neural style transfer.
\newblock \emph{CoRR}, abs/1701.01036, 2017{\natexlab{a}}.
\newblock URL \url{http://arxiv.org/abs/1701.01036}.

\bibitem[Li et~al.(2017{\natexlab{b}})Li, Fang, Yang, Wang, Lu, and
  Yang]{LiFYWL017a}
Li, Yijun, Fang, Chen, Yang, Jimei, Wang, Zhaowen, Lu, Xin, and Yang,
  Ming{-}Hsuan.
\newblock Universal style transfer via feature transforms.
\newblock \emph{CoRR}, abs/1705.08086, 2017{\natexlab{b}}.
\newblock URL \url{http://arxiv.org/abs/1705.08086}.

\bibitem[Li et~al.(2018{\natexlab{b}})Li, Liu, Li, Yang, and
  Kautz]{abs-1802-06474}
Li, Yijun, Liu, Ming{-}Yu, Li, Xueting, Yang, Ming{-}Hsuan, and Kautz, Jan.
\newblock A closed-form solution to photorealistic image stylization.
\newblock \emph{CoRR}, abs/1802.06474, 2018{\natexlab{b}}.
\newblock URL \url{http://arxiv.org/abs/1802.06474}.

\bibitem[Radford et~al.(2021)Radford, Kim, Hallacy, Ramesh, Goh, Agarwal,
  Sastry, Askell, Mishkin, Clark, et~al.]{CLIP}
Radford, Alec, Kim, Jong~Wook, Hallacy, Chris, Ramesh, Aditya, Goh, Gabriel,
  Agarwal, Sandhini, Sastry, Girish, Askell, Amanda, Mishkin, Pamela, Clark,
  Jack, et~al.
\newblock Learning transferable visual models from natural language
  supervision.
\newblock In \emph{International Conference on Machine Learning}, pp.\
  8748--8763. PMLR, 2021.

\bibitem[Rodr{\'\i}guez(2013)]{regular}
Rodr{\'\i}guez, Paul.
\newblock Total variation regularization algorithms for images corrupted with
  different noise models: a review.
\newblock \emph{Journal of Electrical and Computer Engineering}, 2013, 2013.

\bibitem[Shelhamer et~al.(2016)Shelhamer, Long, and Darrell]{FCN}
Shelhamer, Evan, Long, Jonathan, and Darrell, Trevor.
\newblock Fully convolutional networks for semantic segmentation.
\newblock \emph{IEEE transactions on pattern analysis and machine
  intelligence}, 39\penalty0 (4):\penalty0 640--651, 2016.

\bibitem[Wang et~al.(2019)Wang, Fang, Qian, Yang, Zhou, and Zhou]{perspective}
Wang, Ke, Fang, Bin, Qian, Jiye, Yang, Su, Zhou, Xin, and Zhou, Jie.
\newblock Perspective transformation data augmentation for object detection.
\newblock \emph{IEEE Access}, 8:\penalty0 4935--4943, 2019.

\end{thebibliography}

\renewcommand{\thefigure}{A}

\begin{figure*}[htb]
\begin{center}
\includegraphics[width=\linewidth]{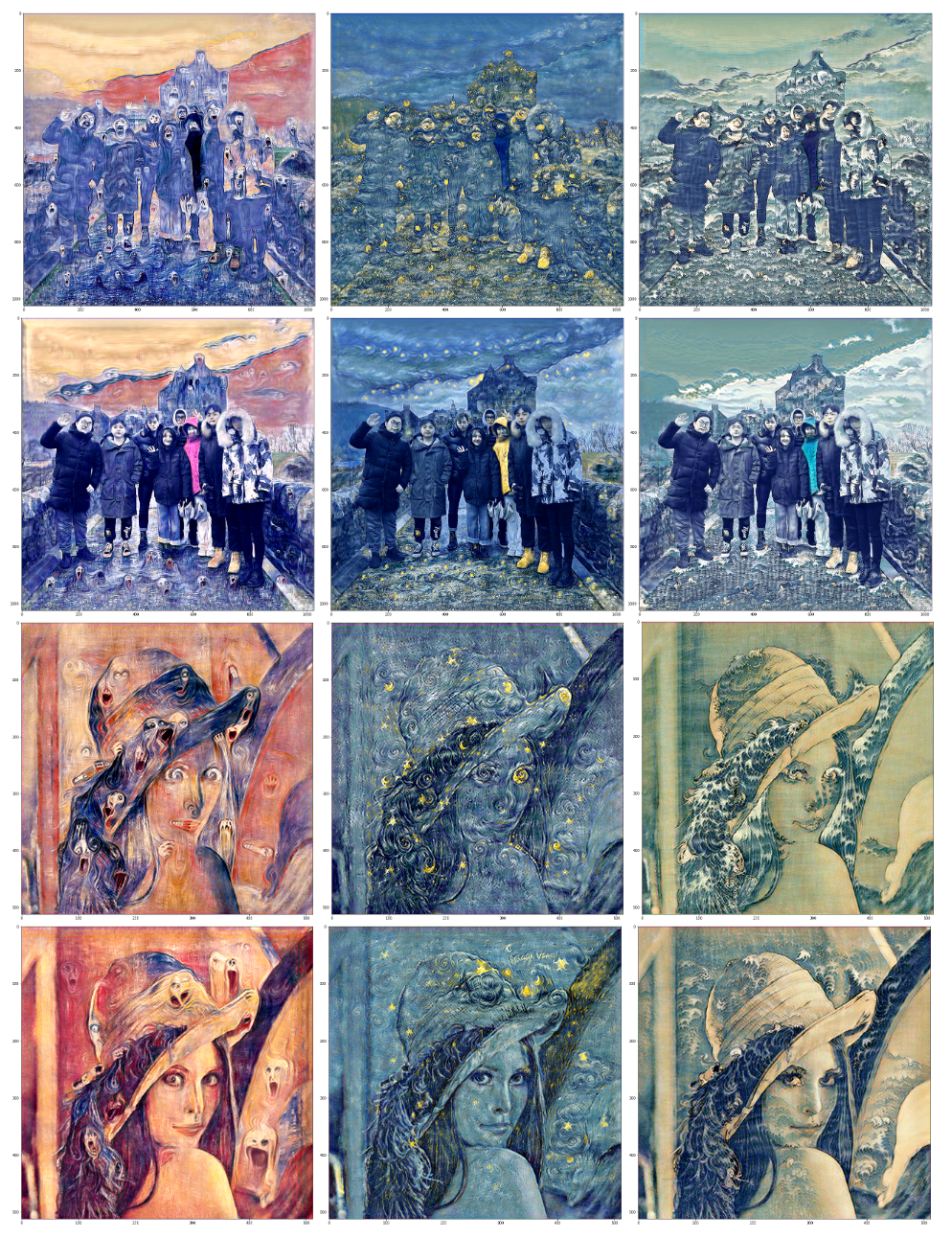}
\end{center}
\caption{A comparison of the generated results of our methods with the original version. The first and third rows are by Basic CLIPstyler, and the second and fourth rows are by Generative Artisan CLIPstyler.}
\label{A comparison of the generated results of our methods with the original version. The first and third rows are by Basic CLIPstyler, and the second and fourth rows are by Generative Artisan CLIPstyler.}
\end{figure*}

\end{document}